%% file: main.tex
\documentclass[letterpaper, 10 pt, conference]{ieeeconf}

\usepackage{00_preamble}
\usepackage{00_macros}
\usepackage{00_spacing}
\IEEEoverridecommandlockouts
\begin{document}

\input{00_title}
\maketitle

\input{0-abstract}
\input{1-introduction}
\input{2-mtt}
\input{3-score-breaks-down}
\input{4-trust-estimation}
\input{5-experiments}
\input{6-conclusion}

\bibliographystyle{IEEEtranS}
\bibliography{references}

\end{document}

%% file: 00_title.tex
\title{\LARGE \bf Bayesian Methods for Trust in Collaborative Multi-Agent Autonomy}


\newif\ifAnonymize

\Anonymizefalse

\ifAnonymize

\else
    \author{R. Spencer Hallyburton and Miroslav Pajic%
    \thanks{This work is sponsored in part by the ONR under agreement N00014-23-1-2206, AFOSR under award number FA9550-19-1-0169, and by the NSF under CNS-1652544 award and the National AI Institute for Edge~Computing Leveraging Next Generation Wireless Networks, Grant CNS-2112562.}
    \thanks{R. S. Hallyburton and M. Pajic are with Department of Electrical and Computer Engineering, Duke University, Durham, NC 27708, USA;
    {\tt\small \{spencer.hallyburton,~miroslav.pajic\}@duke.edu}.}%
    }
\fi

%% file: 0-abstract.tex
\begin{abstract}
    Multi-agent, collaborative sensor fusion is a vital component of a multi-national intelligence toolkit. In safety-critical and/or contested environments, adversaries may infiltrate and compromise a number of agents. We analyze state of the art multi-target tracking algorithms under this compromised agent threat model. We prove that the track existence probability test (``track score'') is significantly vulnerable to even small numbers of adversaries. To add security awareness, we design a trust estimation framework using hierarchical Bayesian updating. Our framework builds beliefs of trust on tracks and agents by mapping sensor measurements to trust pseudomeasurements (PSMs) and incorporating prior trust beliefs in a Bayesian context. In case studies, our trust estimation algorithm accurately estimates the trustworthiness of tracks/agents, subject to observability limitations.
\end{abstract}

%% file: 1-introduction.tex
\section{Introduction}

Networks of low-cost autonomous sensing agents are proliferating in the surveillance and intelligence-gathering space. Sensing networks are used often in \emph{safety-critical, contested environments} such as situational awareness within a national defense strategy. The use of multiple sensors to track dynamic targets in surveillance systems has many benefits including the increased field of view by aggregation and resilience to occlusions, false positives (FPs), and false negatives (FNs) that comes with distributed platforms.

In contested environments, adversaries may infiltrate and compromise one or more agents. Unfortunately, few security analyses have been performed on networks of multiple autonomous agents. There is a great need to analyze classical algorithms for multi-agent collaboration with security in mind. Such analyses are relevant when the collaboration involves untrusted and potentially \emph{distrusted} agents.

We consider a multi-agent \textit{surveillance} problem in which a collection of agents are tasked with jointly observing dynamic objects in a known environment. It is known that optimal data fusion requires fusion of raw detections from each of the platforms in a centralized manner~\cite{1986blackmanRadar}. We consider the classical approach of multi-sensor, multi-target tracking (MTT) using centralized data fusion with a global nearest neighbors data association and Kalman filter state estimator according to~\cite{1986blackmanRadar} and as described in Fig.~\ref{fig:trackers}.

We evaluate the MTT approach under a \emph{compromised agent} threat model. We consider at least one \emph{adversarially compromised} agent provides time-correlated FPs and/or FNs. We then test \emph{the only built-in method of integrity in multi-target tracking}: the ``track score''. The track score is meant to filter FPs by assigning low scores to nascent tracks and to be robust to intermittent FNs; however, it was not designed with security in mind. We prove that \emph{even when benign agents outnumber adversaries}, attackers need only a small number of frames to establish high-confidence FP tracks that are mistakenly believed to be real objects.

Several works have proposed algorithms for ``secure state estimation''. Track score shortcomings were first noted in~\cite{hallyburton2023securing} and a minor modification to the score function was proposed.
\cite{mo2014secure, ren2020secure} designed secure state estimation algorithms for Byzantine attacks on sensors.
\cite{hadjicostis2022trustworthy} considered ``consensus'' in the presence of malicious nodes in distributed estimation.
\cite{mao2022decentralized} derived a distributed and provably-secure state estimation protocol for tracking a dynamical system.

Our approach is orthogonal to secure state estimation and directly estimates whether tracks and agents are \emph{trustworthy} via \emph{trust estimation}. As in Fig.~\ref{fig:trackers}, trust estimation is complementary to (secure) state estimation and can be performed in parallel. Related to trust estimation,
~\cite{wang2006autonomous} explored statistical models of trust assuming binary inputs.
\cite{golle2004detecting} considered vehicular ad hoc networks (VANETs) using a distributed, single-frame trust model to compute agent-based metrics.
\cite{bissmeyer2012assessment} applied a particle filter to track trust and confidence as an ``opinion'' in VANETs assuming certain measurements.
\cite{theodorakopoulos2004trust} used graph theory to extend Dijkstra's shortest path algorithm to trust in ad hoc networks.

There are several shortcomings with existing approaches to trust estimation. First, several works use either binary inputs or single-frame representations of trust (\cite{wang2006autonomous, bissmeyer2012assessment, theodorakopoulos2004trust}). This does not allow for dynamically changing, real-valued, and uncertain outcomes. Furthermore, few works can incorporate prior information into the trust model (\cite{golle2004detecting, theodorakopoulos2004trust}). Prior information is essential in practice with small numbers of agents and imperfect measurements. 

The lack of security awareness in MTT and the inability of existing trust models to capture prior information and uncertainties motivates our novel approach to trust estimation. We formulate the trust estimation problem in a collaborative, multi-agent scenario within the context of Bayesian parameter estimation. In the Bayesian context, a-priori information is incorporated if available via informative priors on agent and track trust parameters. At each timestep, sensor measurements are used by MTT to establish tracks and estimate their states. Our trust model also uses sensor measurements to update the belief on the trustworthiness of those tracks \emph{and} on the agents.

To estimate trust from sensor measurements, we design novel functions that map uncertain sensor data to real-valued ``pseudomeasurements'' (PSMs) of trust. Trust PSMs are reals on $[0,1]$ that are ad-hoc estimates of the trust from a single frame of sensor data. We use PSMs to update the track and agent trust in an alternating procedure inspired by conditional Gibb's sampling. With parametric trust priors and a simple PSM likelihood function, the trust estimation framework performs Bayesian updating of the trust distribution parameters with closed-form, analytic equations. 

We illustrate the effectiveness of our approach to trust estimation in two distinct case studies both with and without prior information. We consider the compromised agent threat model and task trust estimation with ascertaining distributions over the trust of all tracks and agents. Trust estimation successfully verifies tracks on true objects as trusted and tracks on false objects as untrusted under favorable observability conditions. Moreover, we importantly find that prior information is highly useful in accelerating the determination of trust/distrust in multi-agent collaboration.

This paper is organized as follows: Sec.~\ref{sec:MTT} describes the foundations of multiple target tracking, Sec.~\ref{sec:security} proves the vulnerability of classical MTT algorithms, Sec.~\ref{sec:trust-fusion} formalizes our approach to trusted sensor fusion and~\ref{sec:experiments} presents experimental results on trust estimation case studies.

\input{figures/trackers-with-trust}

%% file: figures/trackers-with-trust.tex
\begin{figure}[t!]
    \centering
    \includegraphics[width=0.6\linewidth]{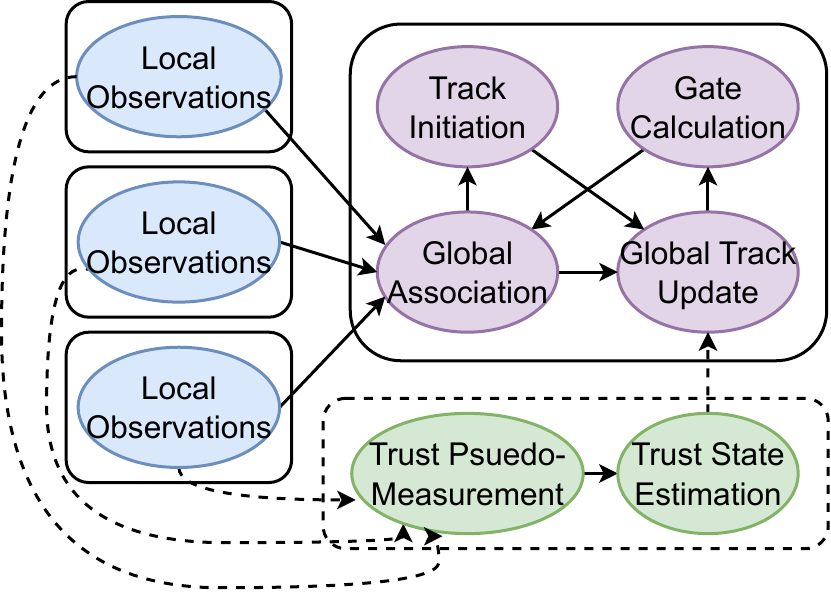}
    \caption{Trust estimation (green) is complementary to existing sensor fusion architectures (purple) for performing inference on data from multiple platforms (blue).}
    \label{fig:trackers}
\end{figure}

%% file: 2-mtt.tex
\section{Multiple Target Tracking (MTT)} \label{sec:MTT}

We consider a multi-target surveillance application where $K$ agents are connected to a centralized data fusion engine and share detection-related data. The central tasks of a multi-target surveillance system is to determine the number of objects that exist in a dynamic scene and to estimate the states of those objects over time; such tasks are known collectively as multiple target tracking (MTT).

\input{figures/surveillance-problem}

False positives (FPs) and missed detections of true objects, i.e., false negatives (FNs), create challenges in the object existence determination task. Fig.~\ref{fig:surveillance-problem} illustrates a common case of ambiguity in MTT regarding existence: one agent believes to see an object that is not seen by the other agents. Naturally, questions arise as to whether this object is real or an FP.
In the following, we present the de facto standard approach to MTT (following e.g.,~\cite{1986blackmanRadar}): a two-step algorithm for solving existence and estimation tasks. We then formally present ``track scoring'' as the classical approach to the existence determination step.

\subsection{MTT as a Two-Step Problem}
We assume that agents $k=1,...,K$ provide $Q_{k,t} \geq 0$ detections at time $t$. We use $Z_t\coloneqq\{z_{q,k,t}\}|_{q\in[1,...,Q_{k,t}],k=1,...,K}$ as the set of the \emph{set of detections} from all agents and $X_t\coloneqq \{x_{i,t}\}$ as the set of all $N_t$ true object states. Formally, the objective of MTT is to estimate the joint posterior:
\begin{align}
    \Pr(X_t | Z_{1:t}) = \frac{\Pr(Z_t|X_t) \Pr(X_t|Z_{1:t-1})}{\Pr(Z_t | Z_{1:t-1})},
\end{align}
where $Pr$ is a probability distribution. At each step, MTT retains a set of tracks, $\hat{X}_t \coloneqq \{\hat{x}_{j,t}\}$ as estimates of object states. Subscripts $j$ do not necessarily align with $i$ since $\hat{X_t}$ estimates both existence and state, e.g.,~$\hat{X}_t$ can have natural FPs or FNs and both $\hat{X}_t,\, X_t$ are permutation-invariant. 

MTT usually takes a two-stage approach to reduce the multi-object posterior to multiple single-object problems. Instead of using all measurements to update all tracks, MTT often assigns measurements to specific tracks for single-track updating (see many examples in~\cite{1986blackmanRadar,bar1995multitarget}). Steps include:
\begin{enumerate}
    \item \textbf{Data association:} perform bipartite matching to assign current detections, $Z_t$, to estimated track states, $\hat{X}_{t-1}$. Often, a measurement can only be used for a single track. Detections without a track start new tracks, tracks without detections are considered ``missed''.
    \item \textbf{Existence \& state estimation:} for each track, use assigned measurements from data association to update the track existence probability and state estimate.
\end{enumerate}

The measurements help the existence task reason about whether the track represents a real object or is an FP . The state estimation task employs an estimator such as the Kalman filter to mix measurements and kinematic models. Important to MTT is both agent pose (i.e.,~position and orientation) and the field of view (FOV) model, $\Phi_k(\cdot)$, that takes as input a point in space and determines if agent $k$ could reasonably observe an object at that point, if there existed one. The FOV model is important e.g.,~so as not to penalize agents and tracks for ``misses' when the candidate track was not in the field of view of the agent to begin with. We group both under the term ``agent characteristics'', $A_t\coloneqq \{a_{k,t}\}$, and assume $A_t$ is known and uncompromised.

\subsection{Track Existence Determination via Likelihood Scoring}

A classic approach to determining whether a track represents a real object or is an FP is to ``confirm'' tracks when they have received a significant number of quality measurements~\cite{1986blackmanRadar,bar1995multitarget}. Confirmation is quantified in a process known as \emph{track scoring} first formalized by~\cite{sittler1964optimal}. It uses hypothesis testing \emph{for each track} as either real ($\mathcal{H}_1$) or fake ($\mathcal{H}_0$). We adopt the notation of~\cite{1986blackmanRadar} that represents the likelihood ratio between the hypotheses~as
\begin{align}
    LR(\hat{x}_{j,t}) = \frac{\Pr(Z_t|\mathcal{H}_1) \Pr(\mathcal{H}_1)}{\Pr(Z_t|\mathcal{H}_0) \Pr(\mathcal{H}_0)} \coloneqq \frac{P_T}{P_F},
\end{align}
where $Z_t$ is the measurement data and $\Pr(\mathcal{H}_i)$ is the prior probability of the hypotheses. The joint distribution of the data and $\mathcal{H}_i$ have probabilities $P_T$ and $P_F$, respectively.

The likelihood ratio evaluated under the natural logarithm is known as the ``track score''. There is a direct transformation between score and the real-object ($\mathcal{H}_1$) probability
\begin{align}
    LLR \coloneqq L = \log \frac{P_T}{P_F}, \quad
    P_T = \frac{e^{L}}{1 + e^{L}} \stepcounter{equation} \tag{{\theequation}a,b}.
\end{align}
The initial track score is set to be
\begin{align} \label{eq:llr-init}
    L_0 = \log \left[ \frac{P_D \beta_{NT}}{\beta_{FP}} \right] 
\end{align}
where $\beta_{NT},\, \beta_{FP}$ are the expected densities of new targets and FPs, respectively. As derived in~\cite{sittler1964optimal}, temporal updates to the track score can be made with the recursion
\begin{align} \label{eq:llr-update}
    L_t &= L_{t-1} + \Delta L_t  \stepcounter{equation} \tag{{\theequation}a} \\ 
    \Delta L_t &= \begin{cases}
        \Delta L_{m,t} \quad \text{if no assignment (miss)} \\
        \Delta L_{h,t} \quad \text{if assignment (hit)}
    \end{cases} \tag{{\theequation}b}\\
    \Delta L_{m,t} &= \log 1-P_D \tag{{\theequation}c} \\
    \Delta L_{h,t} &= \log \left[\frac{P_D}{(2\pi)^{\eta/2} \beta_{FP} \sqrt{|S|}} \right] - \frac{d^2}{2} \tag{{\theequation}d}
\end{align}
where $P_D$ is the probability of detecting a true object, $\eta$ is the number of dimensions, $|S|$ is the determinant of the innovation covariance from the Kalman filter, and $d^2=\Tilde{y}^T S^{-1} \Tilde{y}$ where $\Tilde{y}$ is the innovation in the Kalman filter. A higher $P_D$ yields a greater penalty for a ``miss''. A ``hit'' updates the score as a function of how closely the measurement matches the track's last estimated state.

Track scoring is fundamental to the existence task and the only statistical determination of whether a track represents a real object. Formally, following~\cite{1986blackmanRadar}, track status is:
\begin{equation}
  \begin{aligned}
    \text{status} = 
    \begin{cases}
        \text{track confirmed}; \quad & L \geq \mathcal{T}_2  \\
        \text{continue test}; \quad & \mathcal{T}_1 < L < \mathcal{T}_2 \\
        \text{delete track}; \quad & L \leq \mathcal{T}_1.
    \end{cases}
\end{aligned}
\end{equation}
\begin{align} \label{eq:score-thresholds}
    \mathcal{T}_2 = \log \left[\frac{1-\beta}{\alpha}\right],\quad \mathcal{T}_1 = \log\left[\frac{\beta}{1-\alpha}\right]
\end{align}
with $(\alpha,\, \beta)$ application-specific (see~\cite{1986blackmanRadar, bar1995multitarget}). 

In what follows, we perform a security analysis of track scoring. We show that even under a threat model when benign agents outnumber adversaries, track scoring is vulnerable. This motivates the development of a novel technique in Sec.~\ref{sec:trust-fusion} that quantifies the \emph{trust} of tracks and agents.

%% file: figures/surveillance-problem.tex
\begin{figure}[t!]
    \centering
    \begin{tikzpicture}
        \node (I) {\includegraphics[width=0.8\linewidth]{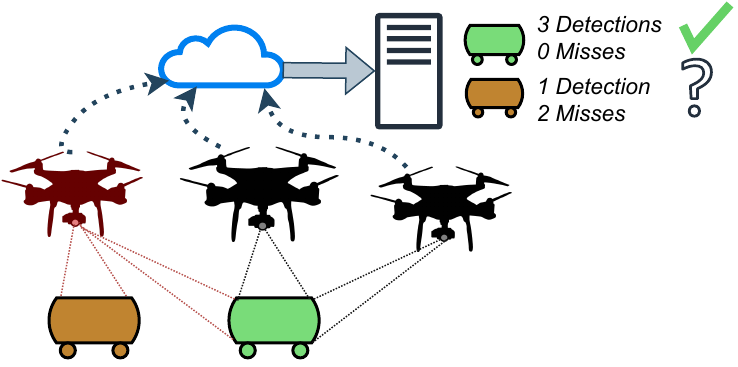}};
        \node [below right=1.28cm and 0.7 of I.west,anchor=west] (O10) {O1};
        \node [right=1.65cm of O10.west,anchor=west] (O20) {O2};
        \node [below left=1.1cm and 3cm of I.east,anchor=west] (T1) {$\bullet$ \emph{Is O1 (yellow) real?}};
        \node [below=0.5cm of T1.west,anchor=west] (T2) {$\bullet$ \emph{Is red agent malicious?}};
    \end{tikzpicture}
    \caption{Consider that a malicious agent (left, red) provides an FP (O1) that is not detected by the benign agents (black). Under what conditions can MTT identify that O1 is an FP? That the red agent is malicious? ``Track scoring'' is a natural tool for existence determination, however, we show it is vulnerable to many adversarial cases. Instead, we propose to augment MTT by estimating the ``trust'' of tracks and agents.}
    \label{fig:surveillance-problem}
\end{figure}
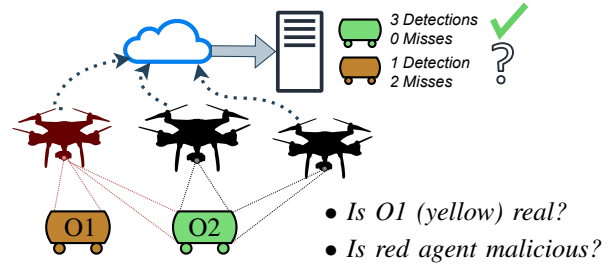

%% file: 3-score-breaks-down.tex
\section{Security Analysis of Track Scoring} \label{sec:security}

The track score represents the belief that a track corresponds to a real object vs. an FP. To motivate a formal analysis, Fig.~\ref{fig:scoring-problem} dives into track scoring for the surveillance problem from Fig.~\ref{fig:surveillance-problem}. O2 (green) is detected by all three agents and will receive a gain contribution from each. On the other hand, O1 (yellow) receives a gain contribution from one agent and two losses from the misses, assuming O1 is within the FOVs of all agents. Despite having more misses than detections, it is not obvious whether O1 will be confirmed; the result depends on the size of the gain/loss contributions.

\input{figures/scoring-problem}

Despite its statistical motivation, in what follows we find that track scoring is vulnerable to adversarial manipulation. We consider a simple threat model and illustrate that even small numbers of adversaries relative to the number of benign agents can quickly lead to incorrect confirmation of an FP. The vulnerability arises because adversaries can create gains that outmatch the losses from benign agents.

\subsection{Threat Model}
We consider $K_a$ adversaries and $K_b$ benign agents. The adversaries can provide fictitious detections of their choosing (FPs) and/or can omit detections of objects within their FOVs (FNs). Adversaries have no way to manipulate the data from benign agents. We assume constant data rates -- i.e., adversaries cannot send data faster or slower than~benign~agents. 

\subsection{Analysis of Track Score Updates}
To perform analysis of MTT in potentially adversarial scenarios, we first bound the change in track score between frames. We then consider the threat model and a scenario in which an adversary wishes to ``confirm'' an FP. We approximate that all detections of the FP are from the adversary and all misses are from benign agents to obtain a more mathematically convenient (yet~suboptimal)~form. 

\subsubsection{Bounding Track Score Gain}
The track score gain depends on the characteristics (noise, deviation from model) of the measurement. Since an adversary can completely control the measurement of an FP, he can achieve any gain up to a maximum fixed by the sensor characteristics and statistical models; these are set a-priori. Prop.~\ref{prop:llr-max} places a bound on the maximum possible gain to the track score for a single frame using these a-priori quantities.

\begin{proposition} \label{prop:llr-max}
    The contribution of any detection to the track score is bounded (from above) by:
    \begin{align*}
        \Delta L_h \leq \log\left[ \frac{P_D}{(2\pi)^{\eta/2}\beta_{FP}\sqrt{|R|}} .\right]
    \end{align*}
\end{proposition}

\begin{proof}
It holds that $S \coloneqq H P H^T + R$ for Kalman filtering where $H$ is the linearization matrix, $P$ the state covariance, and $R$ the measurement covariance;  $\{H P H^T, R\} \geq 0$ by construction, and
    \begin{align*}
        |R| &\leq |HPH^T + R| = |S|
    \end{align*}
    since $\det(A+B)\geq\det(A) + \det(B)$ for positive semi-definite matrices; see e.g.,~\cite{marcus1992survey} for proof. The contribution of a detection hit to the log likelihood is given in~\eqref{eq:llr-update} and without the $-d^2/2$, the contribution is bounded by
    \begin{align*}
        \Delta L_h
        \leq \log \left[\frac{P_D}{(2\pi)^{\eta/2} \beta_{FP} \sqrt{|S|}} \right]
        \leq \log \left[\frac{P_D}{(2\pi)^{\eta/2} \beta_{FP} \sqrt{|R|}} \right]
    \end{align*}
    since $|R| \leq |S|$ and $\log$ is a strictly increasing function.
\end{proof}

\subsubsection{Bounding Change in Track Score}
Now, to bound the total change in track score, we consider the case of $D_t$ detections and $M_t$ misses in Prop.~\ref{prop:track-score-change}.
\begin{proposition} \label{prop:track-score-change}
    For $D_t$ detections and $M_t$ misses, the change in track score in a single frame is bounded by
    \begin{align*}
        \Delta L_t &\leq D_t \log\left[ \frac{P_D}{(2\pi)^{\eta/2}\beta_{FP}\sqrt{|R|}} \right] + M_t \log\left[1-P_D\right]
    \end{align*}
\end{proposition}
\begin{proof}
    The aggregation of misses and hits results in
    \begin{align*}
        \Delta L_t &= M_t\Delta L_{m,t} + D_t\Delta L_{h,t} \\
        \Delta L_{m,t} &= \log\left[1-P_D\right]\\
        \Delta L_{h,t} &\leq \log\left[ \frac{P_D}{(2\pi)^{\eta/2}\beta_{FP}\sqrt{|R|}} \right]
    \end{align*}
    using Prop.~\ref{prop:llr-max}, which concludes the proof. 
\end{proof}

\subsubsection{Natural False Positive Gate Probability is Small}
All sensors exhibit noise, so any volume in the environment can naturally contain FPs. Each volume is statistically independent and the FP density is modeled as a constant, $\beta_{FP}$. The widely-used convention~\cite{1986blackmanRadar,bar1995multitarget} is to model the number of FPs in a bounded volume $V_C$ as a homogeneous Poisson point process:
\begin{align}
    f(N_{FP}=n;\, \Lambda) = \frac{\Lambda^n e^{-\Lambda}}{n!}
\end{align}
with $N_{FP}$ a number of FPs and $\Lambda = V_C \beta_{FP}$.

Measurements are assigned to tracks in Step 1 of MTT if they satisfy the \emph{gating} criteria. Simply put, a measurement is allowed to update a track if they are statistically ``close to'' each other (i.e., if the measurement is within the ``gating volume'' of the track). It is possible for a natural FP to be close to an established track and satisfy the gating criteria; we consider the probability of this occurrence in Prop.~\ref{prop:fp-det-prob}.

\begin{proposition} \label{prop:fp-det-prob}
    In an environment with constant FP density $\beta_{FP}$, at least one natural FP will be within the gating volume $V_G$ of an existing track with probability $1 - e^{-V_G \beta_{FP}}$.
\end{proposition}
\begin{proof}
    Suppose a confirmed track exists and on a round of measurements the volume of its gating region is $V_G$. Then, 
    \begin{align*}
        \Pr(N_{FP} \geq 1 \in V_G) &= 1 - \Pr(N_{FP}=0 \in V_G)\\
        &= 1 - F(N_{FP}=0;\, \Lambda_G)\\
        &= 1 - e^{-\Lambda_G} = 1 - e^{-V_G \beta_{FP}},
    \end{align*}
concluding the proof.
\end{proof}

A target tracking scenario might have $\beta_{FP}=\num{e-6}$ and $V_G(M) \propto \sqrt{|S|}$ (see~\cite{1986blackmanRadar} for full definition of $V_G$), thus making the probability of FP gate for a track small. For illustration, assuming independent FPs across frames, we might observe only 1 gate of a benign FP in a volume element on 100 seconds of data at 10 Hz data rates.

\subsubsection{Track Score Under Threat Model}
Finally, we consider that the adversary wishes for MTT to believe an FP is a real object. Specifically, we assume the adversary wishes to provide false detections to achieve FP confirmation via the track score as quickly as possible in the presence of benign agents that are providing negative results (i.e.,~no detections). Theorem~\ref{thm:k-frames-to-confirm} asserts the minimum number of frames, $T_{min}$, after which the score of an FP is above the confirmation threshold (i.e., an FP is confirmed). 

\begin{theorem}\label{thm:k-frames-to-confirm}
    Given $K_a$ adversaries and $K_b$ benign agents observing a single volume element, an adversary can establish a valid track in a minimum of $T_{min}$ steps, where
    \begin{align*}
        T_{min} &\approx 1 + \frac{
        \mathcal{T}_2 - \log \left[ \frac{P_D \beta_{NT}}{\beta_{FP}} \right] 
        }{
        \left[ K_a \log\left[ \frac{P_D}{(2\pi)^{\eta/2}\beta_{FP}\sqrt{|R|}} \right] + K_b \log\left[1-P_D\right] \right]
        }
    \end{align*}
    frames, with $\mathcal{T}_2$ set according to~\eqref{eq:score-thresholds}.
\end{theorem}
\begin{proof}
    A track is confirmed if $L_t \geq \mathcal{T}_2$. Also, $L_t=L_{t-1}+\Delta L_t$ so $L_T = L_0 + \sum_{t=1}^T \Delta L_{t}$. Prop.~\ref{prop:track-score-change} bounds the change in track score for any $N_t$ detections and $M_t$ misses. Since the track is an adversarial FP, $N_t$ will be a combination of one adversarial FP from the malicious agent and some number of natural FPs from benign agents that coincidentally overlap. Similarly, $M_t$ will be one miss from each benign agent not providing a natural FP, i.e.,~$K_a \leq N_t \leq K_a + K_b$ and $0 \leq M_t \leq K_b$. However, by Prop~\ref{prop:fp-det-prob}, the probability of benign agents having natural FPs near the adversary's FP will be sufficiently small such that $N_t$ is nearly completely determined by the adversaries and $M_t$ is made up of all benign agents that can observe the candidate, i.e.,~$N_t\approx K_a$ and $M_t\approx K_b$. Since there are no frame-dependent terms in $\Delta L_t$, after transforming the sum to get the threshold point, using~\eqref{eq:llr-init} for initial score, we obtain
    \begin{align*}
        (T_{min}&-1) \Delta L_t = \mathcal{T}_2 - \log \left[ \frac{P_D \beta_{NT}}{\beta_{FP}} \right] \\
        T_{min} &\approx 1 + \frac{
        \mathcal{T}_2 - \log \left[ \frac{P_D \beta_{NT}}{\beta_{FP}} \right] 
        }{
        \left[ K_a \log\left[ \frac{P_D}{(2\pi)^{\eta/2}\beta_{FP}\sqrt{|R|}} \right] + K_b \log\left[1-P_D\right] \right]
        }.
    \end{align*}
\end{proof}

\input{figures/frames-to-confirm}

Theorem~\ref{thm:k-frames-to-confirm} establishes a minimum number of frames after which MTT will erroneously confirm an FP as a valid object. This assumes the adversary places detections to optimally increase gain but requires no prior information about the environment. Figure~\ref{fig:frames-to-confirm} shows a surface plot of the number of frames as a function of $(K_a,\,K_b)$ while fixing parameters to nominal values. Importantly, \emph{even when the number of benign agents outnumbers the adversary, the adversary easily achieves a confirmed track in a matter of single-digit frames.} For example, when $K_a=1$ and $K_b=3$, adversaries only require 6 frames at minimum to confirm an FP.

%% file: figures/scoring-problem.tex
\begin{figure}[t!]
    \centering
    \begin{tikzpicture}
        \node (I) {\includegraphics[width=0.8\linewidth]{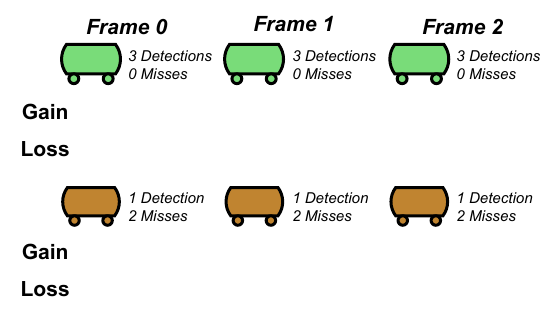}};
        \scriptsize  
        \node [above right=1.23cm and 1.07 of I.west,anchor=west] (O20) {O2};
        \node [right=2.07cm of O20.west,anchor=west] (O21) {O2};
        \node [right=2.07cm of O21.west,anchor=west] (O22) {O2};
        \node [below right=0.56cm and 1.07 of I.west,anchor=west] (O10) {O1};
        \node [right=2.07cm of O10.west,anchor=west] (O11) {O1};
        \node [right=2.07cm of O11.west,anchor=west] (O12) {O1};
        \node [above right=0.53cm and 1cm of I.west,anchor=west] (10G) {$\sum_i^3 \Delta L^i_{h,0}$};
        \node [right=2cm of 10G.west,anchor=west] (11G) {$\sum_i^3 \Delta L^i_{h,1}$};
        \node [right=2cm of 11G.west,anchor=west] (12G) {$\sum_i^3 \Delta L^i_{h,2}$};
        \node [below right=0.42cm and 0.5cm of 10G.west,anchor=west] (10L) {0};
        \node [right=2cm of 10L.west,anchor=west] (11L) {0};
        \node [right=2cm of 11L.west,anchor=west] (12L) {0};
        \node [below right=1.75cm and 0.2cm of 10G.west,anchor=west] (20G) {$\Delta L^1_{h,0}$};
        \node [right=2cm of 20G.west,anchor=west] (21G) {$\Delta L^1_{h,1}$};
        \node [right=2cm of 21G.west,anchor=west] (22G) {$\Delta L^1_{h,2}$};
        \node [below=0.42cm of 20G.west,anchor=west] (20L) {$\sum_i^2 \Delta L^i_{m,1}$};
        \node [right=2cm of 20L.west,anchor=west] (21L) {$\sum_i^2 \Delta L^i_{m,1}$};
        \node [right=2cm of 21L.west,anchor=west] (22L) {$\sum_i^2 \Delta L^i_{m,2}$};
    \end{tikzpicture}
    \vspace{-10pt}
    \caption{Track scoring represents the probability that an object exists. Increments are calculated using gains from detections and losses from misses. With many detections, the true-object hypothesis score for O2 will increase. With a mix of detections and misses, the outcome for O1 is not obvious. We prove the conditions under which O2 is confirmed despite few detections and consistent misses in Theorem~\ref{thm:k-frames-to-confirm}.}
    \label{fig:scoring-problem}
\end{figure}

%% file: figures/frames-to-confirm.tex
\begin{figure}[t!]
\centering
\begin{tikzpicture}
    \node (I) {\includegraphics[trim={2.5cm .2cm 1.5cm 1.2cm},clip,width=0.58\linewidth]{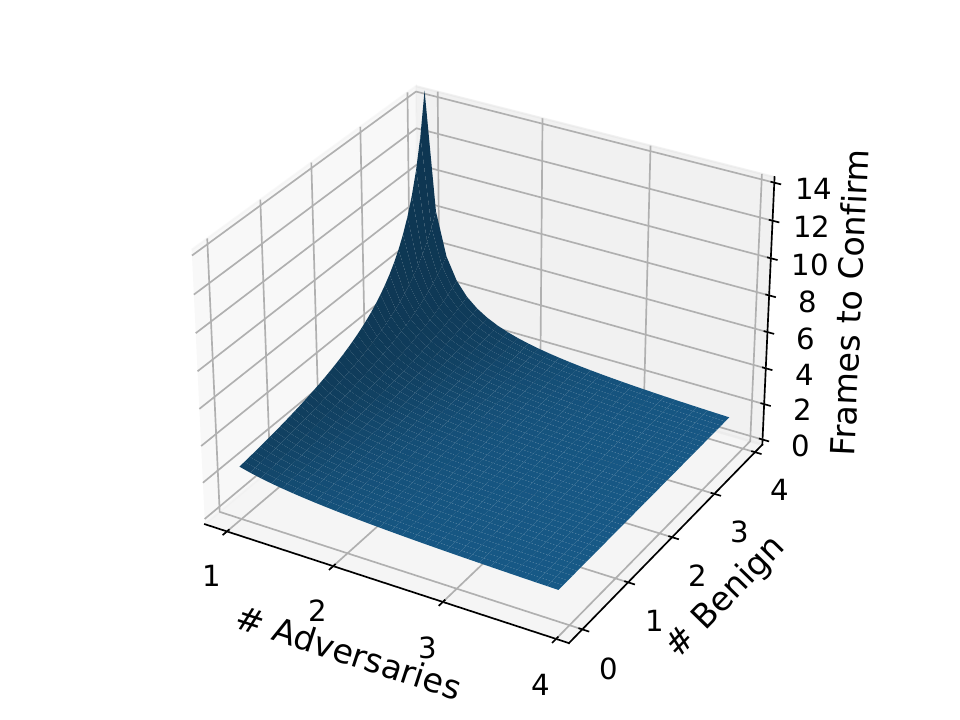}};
    \node [above right=0.2cm and 0.07cm of I.east,anchor=west] (A) at (2.75, 1.50)  {\underline{Parameters:}};
    \node [below=0.5cm of A.west,anchor=west]  (B) {$\bullet \ P_D=0.9$};
    \node [below=0.5cm of B.west,anchor=west]  (C) {$\bullet \ \beta_{FP}=\num{e-6}$};
    \node [below=0.5cm of C.west,anchor=west]  (D) {$\bullet \ \beta_{NT}=\num{e-9}$};
    \node [below=0.5cm of D.west,anchor=west]  (E) {$\bullet \ |R|=5$};
    \node [below=0.5cm of E.west,anchor=west]  (F) {$\bullet \ M=3$};
    \node [below=0.5cm of F.west,anchor=west]  (G) {$\bullet \ T_2=\log\left[\frac{1-\beta}{\alpha}\right]$};
    \node [below=0.5cm of G.west,anchor=west]  (H) {$\bullet \ \alpha=\num{e-6},\, \beta=\num{e-2}$};
\end{tikzpicture}
\caption{Following Theorem~\ref{thm:k-frames-to-confirm}, even with only few agents viewing an object, adversaries can quickly confirm fake tracks. More benign agents viewing a track forces adversaries to use more time to establish a confirmed track. E.g., evaluating $f(K_a=1,K_b=1)\approx3$,  $f(K_a=1,K_b=3)\approx6$.}
\label{fig:frames-to-confirm}
\end{figure}

%% file: 4-trust-estimation.tex
\section{Estimation of Track and Agent Trust in MTT} \label{sec:trust-fusion}

As shown in Sec.~\ref{sec:security}, MTT lacks the security-awareness needed to correctly identify an FP even when benign agents outnumber adversarial agents. To overcome this vulnerability, we consider that agents' detections can inform whether the agents and the tracks they help establish are \emph{trusted}. Trust can then be used to inform MTT to ignore distrusted tracks.

Prior works have considered secure state estimation~\cite{mo2014secure, ren2020secure, hadjicostis2022trustworthy, mao2022decentralized} or trusted multi-agent collaboration~\cite{wang2006autonomous, golle2004detecting, bissmeyer2012assessment, theodorakopoulos2004trust}. Ours is the first to estimate the trustworthiness of tracks and agents within an MTT context. Informally, trust is quantified as a belief over $[0, 1]$ of a track existing (track trust) or of agents providing measurements consistent with the true state of the world (agent trust).
To support the derivation of trust estimation algorithms and for case study in Sec.~\ref{sec:experiments}, we present two cases of multi-agent MTT in Fig.~\ref{fig:trust-cases}. In both cases, three agents are providing detections from partially overlapping FOVs. Detections are fed to the central MTT that establishes global tracks. In Case 1, Agents 0 and 2 are both trying to establish malicious FPs (Tracks 2 and 5). In Case 2, agent 2 is trying to establish two malicious FPs (Tracks 4, 5). We propose and evaluate algorithms that leverage these discrepancies to estimate the trustworthiness of each track and each agent.

\input{figures/trust-cases}
\subsection{MTT With Trust Estimation Posterior}
Formally, the MTT with trust estimation objective is to estimate the full joint posterior:
\begin{equation}\label{eq:trust-posterior}
\begin{aligned}
    \Pr(X_{t},& \Tau^c_{t}, \Tau^a_{t} | Z_{1:t}, A_{1:t}) \\
     &= \Pr(\Tau^c_{t}, \Tau^a_{t} | Z_{1:t}, A_{1:t}) \Pr(X_{t} | \Tau^c_{t}, \Tau^a_{t}, Z_{1:t}, A_{1:t}) 
\end{aligned}
\end{equation}
where $X_{t}$ are object states for all $N$ objects, $\Tau^c_{t}$ are track trusts for each $j=1...C$ tracks, $\Tau^a_{t}$ are agent trusts for each $k=1...K$ agents, $Z_{1:t}$ are measurements from all agents, and $A_{1:t}$ are agent characteristics including the pose and FOV model, $\Phi_k(\cdot)$. In~\eqref{eq:trust-posterior}, we use conditional probability to decompose into subproblems: (\ref{eq:trust-posterior}.1) trust estimation posterior, (\ref{eq:trust-posterior}.2) state estimation posterior conditioned on trust. This decomposition allows us to run trust estimation as its own node independent of MTT, as illustrated in Fig.~\ref{fig:trackers}. The remainder of this works is concerned with (\ref{eq:trust-posterior}.1), the trust estimation posterior. We leave a full treatment of (\ref{eq:trust-posterior}.2), target tracking augmented with trust, to future works.

To estimate the trust posterior, we use a decomposition inspired by the popular Gibbs sampling (see e.g.,~\cite{shemyakin2017copula}). This breaks the trust posterior of (\ref{eq:trust-posterior}.1) into an alternating two-step process leveraging conditional probabilities, i.e., 
\begin{equation}\label{eq:trust-est-posterior}
\begin{aligned}
    (1)\ &\text{Update track trust:} \ \Pr(\Tau^c_{t}\ |\ \Tau^a_{t-1}, Z_{1:t}, A_{1:t}) \\
    (2)\ &\text{Update agent trust:} \ \Pr(\Tau^a_{t}\ |\ \Tau^c_{t}, Z_{1:t}, A_{1:t}).
\end{aligned}
\end{equation}
With this separation, we can update track/agent trusts sequentially. The drawback of a Gibbs-style approach is a loss of formal convergence guarantees for the general case; in our case of simple univariate trust distributions with two parameters, we observe rapid convergence in practice.

\subsection{Trust Pseudomeasurements (PSMs)}
Unfortunately, there is no function to explain the likelihood of the data given the trust, $\Pr(Z_{1:t}|\Tau^a_{t},\Tau^c_{t},A_{1:t})$, making an exact Bayesian approach to trust estimation intractable. Instead, we introduce \emph{trust pseudomeasurements} (PSMs) and the approximations
\begin{align*}
\Pr(\Tau^c_{t} |& \Tau^a_{t-1}, Z_{1:t}, A_{1:t}) \\
&\approx \Pr(\Tau^c_{t} | g^c(\Tau^a_{t-1}, Z_{1:t}, A_{1:t})) = \Pr(\Tau^c_{t} | \Rho^c_{1:t}) \\
\Pr(\Tau^a_{t} |& \Tau^c_{t}, Z_{1:t}, A_{1:t}) \\
&\approx \Pr(\Tau^a_{t} | g^a(\Tau^c_{t}, Z_{1:t}, A_{1:t})) = \Pr(\Tau^a_{t} | \Rho^a_{1:t})
\end{align*}
where $g^c,\, g^a$ denote track/agent-focused PSM functions that map the measurements to the trust domain of $[0,1]$. What follows from this is an \emph{ad-hoc measurement} of track and agent trust at every frame $\Rho^c \leftarrow \{\rho^c_j\}$ and $\Rho^a \leftarrow \{\rho^a_k\}$.

Each PSM is a set of datapoints that each contain a value and an uncertainty; this is akin to e.g.,~a position measurement that provides a measured value along with a standard deviation of the measurement's uncertainty. Each PSM datapoint uses information only from a single track-agent pair, (track $j$, agent $k$). The PSM is then $\rho_j=\{(v_{j,k}, c_{j,k})\}$ where each $(v_{j,k}, c_{j,k})$ is a PSM datapoint, $v_{j,k}\in [0, 1]$ is the datapoint's value, and $c_{j,k}\in [0,1]$ is the confidence (uncertainty) in the datapoint's value. The subscripts $(j,k)$ indicate the datapoint leveraging information from track $j$ and agent $k$. For example, a track $j'$ may receive PSM datapoints from each of the agents such that its PSM is $\rho_{j'}=\{(v_{j',1}, c_{j',1}),(v_{j',2}, c_{j',2}),...\}$.

Notably, not all agents will see all tracks; the expected set of observations on each frame is informed by the FOV model for each agent, $\Phi_k(\cdot)$, which is an indicator function returning \texttt{True} or \texttt{False}. Alg.~\ref{alg:trust-psm-track} presents the PSM routine for a track: each agent expected by the FOV model to see the track provides a PSM datapoint with value of whether or not the agent has a detection near the track and confidence of the agent's trust; the confidence manifests conditional Gibb's sampling. Alg.~\ref{alg:trust-psm-agent} presents the PSM routine for an agent: each track at the central MTT that the agent is expected to see by its FOV model provides a PSM datapoint with value as the expectation of the track trust ($\expectation$) if the agent saw the track or the negation of track trust if the agent did not see the track. Confidence is set by variance ($\variance$) of the track trust; the value and confidence manifest Gibb's sampling.

\input{algorithms/pseudomeasurements}

\subsection{Trust Estimation}

After introducing trust PSMs and making independence assumptions, we have reduced the estimation problem to the posteriors $\Pr(\Tau^c_{j,t} | \Rho^c_{j,1:t})$, $\Pr(\Tau^a_{j,t} | \Rho^a_{j,1:t})$. We assume the PSMs are i.i.d.; this is sub-optimal as the construction of PSMs requires verification against other agents, generating inter-agent correlations. We also expect the PSMs to exhibit autocorrelation due to the temporal nature of tracking. These assumptions, while sub-optimal, enable the use of simple parameter estimation algorithms.

A Bayesian approach is warranted from the perspective of small sample sizes and prior knowledge. Tracking will generate relatively few PSMs since the PSM function requires FOV overlap from multiple agents; observation density is expected to be sparse. Significant prior knowledge may also be available in the form of correlations between platform types or prior trust/distrust of particular agents.

The estimation process is identical for track and agent trust posteriors. The unknown parameters $\theta$ are random variables. The probability distribution of trust for tracks is:
\begin{equation}
\begin{aligned}
    \Pr(\tau^c_j|\Rho^c_j) &= \int \Pr(\tau^c_j,\theta^c_j|\Rho^c_j) d\theta^c_j \\
    &= \int \Pr(\tau^c_j|\theta^c_j,\Rho^c_j) \Pr(\theta^c_j|\Rho^c_j) d\theta^c_j \\
    &= \int \Pr(\tau^c_j|\theta^c_j) \Pr(\theta^c_j|\Rho^c_j) d\theta^c_j.
\end{aligned}
\end{equation}
The parameter posterior is:
\begin{align}
    \Pr(\theta^c_j | \Rho^c_j) & \propto \Pr(\Rho^c_j|\theta^c_j) \Pr(\theta^c_j).
\end{align}
The same procedure applies for agent trust, $\Pr(\tau^a_k|\Rho^a_k)$. 

In practice, tracks are in either the state of being true objects or FPs. Thus, $\tau^c_j$ is the belief of a track being in one state or the other. As such, $\Pr(\Rho^c_j|\theta^c_j)$ naturally takes on a Bernoulli distribution with a single parameter, $\theta^c_j$. For a Bernoulli likelihood, the standard choice of prior $\Pr(\theta^c_j)$ is the Beta distribution~\cite{shemyakin2017copula}. The Beta has two parameters (``hyperparameters''), $(\alpha,\, \beta)$, and is conjugate to the Bernoulli likelihood meaning that the posterior, $\Pr(\theta^c_j | \Rho^c_j)$, is also a Beta distribution. 
It is well-known that the Bayesian parameter update for the Beta-Bernoulli pair has a closed form. As the trust estimation update, we use each of the PSM datapoints to update the posterior parameters. For a track with PSM $\rho_j=\{(v_{j,k}, c_{j,k})\}$, the posterior of the trust parameter distribution, $\Pr(\theta^c_j | \Rho^c_j)$, will be updated via:
\begin{equation}\label{eq:beta-bernoulli-update}
    \begin{aligned}
        \alpha^c_{j,t} &= \alpha^c_{j,t-1} + \sum_k c_{j,k} v_{j,k} \\
        \beta^c_{j,t} &= \beta^c_{j,t-1} + \sum_k c_{j,k} (1-v_{j,k}).
    \end{aligned}
\end{equation}
The same process applies for agent parameters, $(\alpha^a_{k,t}, \beta^a_{k,t})$.

\subsection{Simplified Trust-Aware Sensor Fusion}
The final step is to perform trusted state estimation. A general model would consider that in addition to FPs/FNs the adversary could furnish incorrect state estimates of true objects; e.g., a translation outcome~\cite{2022hally-frustum}. In this work, we limit the adversary to only FP/FN outcomes. Thus, trust estimation is sufficient to confirm or remove tracks from the database.

%% file: figures/trust-cases.tex
\begin{figure}[t!]
    \begin{subfigure}[t]{0.42\linewidth}
    \centering
    \includegraphics[width=\linewidth]{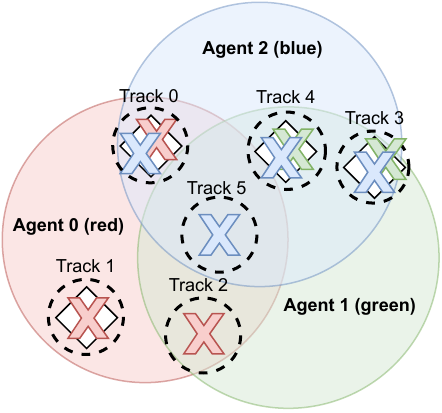}
    \caption{Case 1}
    \label{fig:trust-case-1}
    \end{subfigure}
    %
    \begin{subfigure}[t]{0.42\linewidth}
    \centering
    \includegraphics[width=\linewidth]{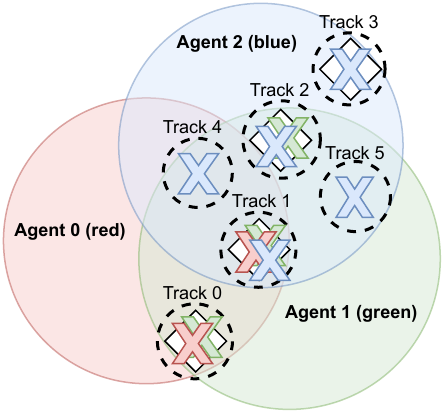}
    \caption{Case 2}
    \label{fig:trust-case-2}
    \end{subfigure}
    %
    %
    \begin{subfigure}[t]{0.13\linewidth}
    \centering
    \includegraphics[width=\linewidth]{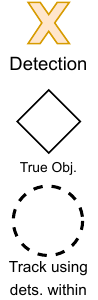}
    \end{subfigure}
    \caption{Three agents with partially overlapping FOVs share detections with MTT to establish tracks. (Case 1) Agents 0 and 2 providing FP detections to try to establish malicious tracks, Tracks 2, 5. (Case 2) Agent 2 providing two FP detections for two malicious tracks, Tracks 4, 5. If any track is in only a single agent's FOV (e.g.,~Case 1, Track 1), not enough information is available to estimate track trust. With multiple overlapping observations (e.g.,~Case 2, Track 1), information from each agent will be used in concert with FOV models to form trust PSMs on tracks and agents.}
    \label{fig:trust-cases}
\end{figure}

%% file: algorithms/pseudomeasurements.tex
\begin{algorithm}
\caption{Trust pseudomeasurement for track $\hat{x}^c_j$}\label{alg:trust-psm-track}
\begin{algorithmic}
\Require $K>0$ agents with trusts $\tau^a_k$, $Z\gets\{z_k\}$ detections from agents, $\hat{x}^c_j$ track of interest, FOV functions $\Phi_k(\cdot)$
\Ensure $\rho^c_j$ if $N_{exp}>1$ else $[\,]$
\State $\rho^c_j \gets [\,]$
\For{$k=1...K$} \Comment{loop over agents}
\If{$\Phi_k(\hat{x}^c_j)$} \Comment{if expected to see}
    \State $N_{exp} \gets N_{exp} + 1$
    \If{$\exists z_{i,k} \in z_k \ \text{s.t.}\ \texttt{dist}(z_{i,k}, \hat{s}^c_j) \ \text{is small}$}
        \State $v_{j,k} \gets 1.0,\, c_{j,k} \gets \tau^a_{k} $
    \Else \Comment{if not observed}
        \State $v_{j,k} \gets 0.0,\, c_{j,k} \gets \tau^a_{k} $
    \EndIf
    \State $\rho^c_j.\texttt{append}((v_{j,k},\, c_{j,k}))$
\EndIf
\EndFor
\end{algorithmic}
\end{algorithm}


\begin{algorithm}
\caption{Trust pseudomeasurement for agent $k$}\label{alg:trust-psm-agent}
\begin{algorithmic}
\Require $\Phi_k(\cdot)$ FOV for agent k, $\hat{X}_{c} \gets \{\hat{x}^c_j\}$ tracks from MTT with trusts $\tau^c_{j}$, $\hat{X}_{k} \gets \{\hat{x}^k_{j'}\}$ tracks from agent $k$'s local estimation.
\Ensure $\rho^a_k$
\State $\rho^a_k \gets [\,]$
\For{$\hat{x}^c_j \in \hat{X}_{c}$} \Comment{loop over tracks from central}
    \If{$\Phi_k(\hat{x}^c_j)$} \Comment{if expected to see}
        \If{$\hat{x}^c_j \in \hat{X}_{k}$} \Comment{if agent has match}
            \State $v_j \gets \expectation[\tau^c_{j}], \, \quad \ \ c_j \gets 1 - \variance[\tau^c_{j}]$
        \Else \Comment{if agent does not have match}
            \State $v_j \gets 1 - \expectation[\tau^c_{j}], \, c_j \gets 1 - \variance[\tau^c_{j}]$
        \EndIf
        \State $\rho^a_k.\texttt{append}((v_j,\, c_j))$
    \EndIf
\EndFor

\end{algorithmic}
\end{algorithm}

%% file: 5-experiments.tex
\section{Multi-Agent Trust Experiments} \label{sec:experiments}

In this section, we evaluate the proposed trust estimation models on two case studies and two sets of prior information. The trust estimation model ingests PSMs that are formed based on data from agents. The PSMs update the parameters of the Beta posteriors. We find the availability of prior information on agent trust influences the certainty with which the model identifies (dis)trusted agents and tracks.

\subsection{Models}

\input{figures/trust-outcomes}

We consider an environment with three static agents and partially overlapping circular FOVs in the 2D plane as in Fig.~\ref{fig:trust-cases}. The agents each make 2D observations of objects. We neglect benign FPs by assuming MTT can filter transient detections given the lack of temporal persistence.
%
We evaluate two cases with adversaries. First, two agents are partially adversarially compromised and are providing malicious data in the form of persistent FPs in the FOV. Second, only a single agent is providing FPs. The cases are described in Figs.~\ref{fig:trust-case-1} and`\ref{fig:trust-case-2}, respectively.

To perform multi-agent, multi-target surveillance, we implement a Kalman-filter-based multi-sensor MTT algorithm with canonical track scoring as the fusion engine according to~\cite{1986blackmanRadar}. We perform Bayesian estimation of the parameter posteriors, $\Pr(\theta^c_j | \Rho_j^c),\, \Pr(\theta^a_k | \Rho_k^a)$, for both track trust and agent trust estimation. We reparameterize the Beta from its canonical $(\alpha, \beta)$ form to a $(\lambda \phi, \lambda (1-\phi))$ form where $\phi = \alpha / (\alpha + \beta) $ is the mean and $ \lambda = \alpha + \beta $ is known as the ``precision''. For each case, we consider two prior conditions. (1) There is no prior information available regarding track/agent trust. In this case, an uninformative prior on all parameters is appropriate; e.g.,~$\theta \sim \text{Beta}(0.5,1)$\footnote{$\text{Beta}(0.5,1)$ has modes near the extrema and reflects that tracks either exist or do not exist and is more uninformative than a uniform prior~\cite{shemyakin2017copula}.}. (2) There is prior information that Agent 1 is trusted. In this case, a prior of~$\theta_{k_1} \sim \text{Beta}(0.8,10)$ is heuristically chosen for Agent 1 and the uninformative prior for all others. 

At each step, we add a small amount of process noise to the trust posteriors by decreasing the precision parameter, $\phi$, for all tracks and all agents. This reflects that, in the absence of measurements, we should become less confident about our trust estimates over time. When PSMs arrive, we use the closed-form Bayesian update formula for the Beta-Bernoulli conjugate pair described in~\eqref{eq:beta-bernoulli-update}.


\subsection{Results}

Trust posteriors for tracks and agents in both cases with two sets of prior information are illustrated in Fig.~\ref{fig:trust-outcomes}. In the following, we describe the observed outcomes for each case.

\paragraph{Case 1, no prior; Figs~\ref{fig:trust-outcomes-track-1-no-prior}, \ref{fig:trust-outcomes-agent-1-no-prior}.} Tracks corresponding to valid objects within multiple agents' FOVs are confirmed trusted (Tracks 0, 3, 4). Track 1 corresponds to a valid object, however, it is only visible in a single agent's FOV (see Fig.~\ref{fig:trust-case-1}). Therefore, no trust update is performed and the track's trust remains as the prior. Track 5 (an FP) is viewable from all agents' FOVs, however since only Agent 2 is detecting it, the model distrusts Track 5. On the other hand, since Track 2 (an FP) is only viewable from Agents 0 and 1, the model cannot resolve whether Agent 0 (detecting track 2) or Agent 1 (not detecting track 2) is the correct outcome. Because of this ambiguity, the model finds it difficult to determine the trust on Agents 0 and 1.

\paragraph{Case 1, prior on Agent 1; Figs~\ref{fig:trust-outcomes-track-1-prior}, \ref{fig:trust-outcomes-agent-1-prior}.} The addition of prior information on the trust of Agent 1 significantly improves the Case 1 outcomes. Tracks corresponding to true objects behave similarly as without the prior. Now, the model disambiguates the discrepancy between Agents 0 and 1 on Track 2; it now believes Agent 0 must be untrustworthy and providing FP detections to track 2 (an FP). The influence of prior information is also reflected in the agent trust posterior as Agent 1 is now believed to be trusted.

\paragraph{Case 2, no prior; Figs~\ref{fig:trust-outcomes-track-2-no-prior}, \ref{fig:trust-outcomes-agent-2-no-prior}.} Tracks corresponding to valid objects in shared regions of the FOVs are trusted (Tracks 0, 1, 2). Tracks isolated in a single agent's FOV maintain the prior (Track 3) As opposed to Case 1, the two FPs in Case 2 both originate from Agent 2. It is then easier for the model to identify that Agent 2 is distrusted since its information is inconsistent with both of the other agents. Consequentially, both FP tracks (Tracks 4, 5) and Agent 2 tend towards distrusted while Agents 0, 1 tend toward trusted.

\paragraph{Case 2, prior on Agent 1; Figs~\ref{fig:trust-outcomes-track-2-prior}, \ref{fig:trust-outcomes-agent-2-prior}.} The addition of prior information in the form of a strong prior on Agent 1 accentuates the trust outcomes. The FP tracks become more distrusted while the trusted agents become more trusted. 

\subsection{Discussion}

Several general observations can be made about the trust model from these select case studies. First, the model cannot verify the trust of tracks that are only visible from a single agent. In these cases, the distribution over trust remains as the prior. Such occurrences could be used as input to a sensor resource management task that can dynamically direct sensing resources to mitigate uncertainty. Second, an even mix of positive (hit) and negative (miss) events for a single track from multiple agents is irresolvable in the absence of prior information. Prior information such as the prior probability of FPs vs FNs or a prior belief on the trust of agents can help to resolve such an ambiguity. Third, accurate prior information significantly improves the model's ability to estimate trust. Due to the alternating conditional Gibb's sampling step, agent trust is used to estimate track trust and vice verse. Thus, having a prior on either agent or track trust makes an immediate impact on the trust estimation process. 

%% file: figures/trust-outcomes.tex
\begin{figure*}[t!]
    \centering
    \begin{subfigure}[t]{0.24\linewidth}
        \centering
        \includegraphics[width=\linewidth]{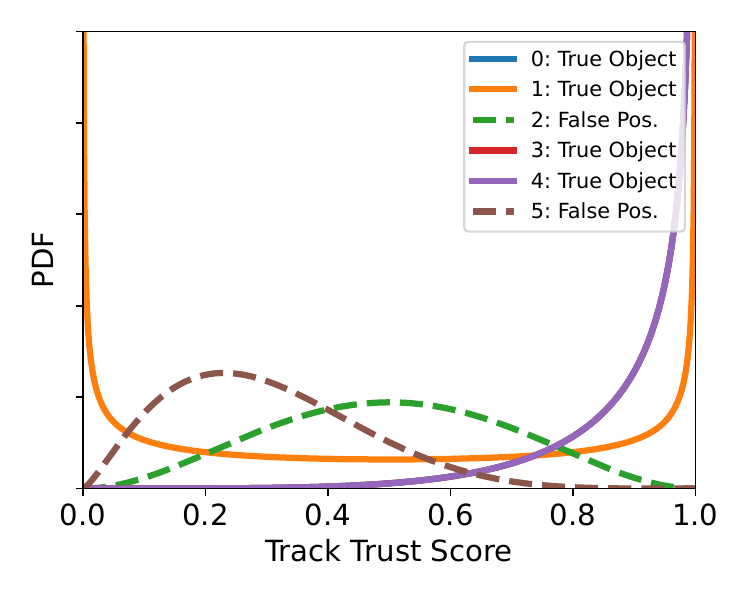}
        \caption{C1: track trust, uninf. prior}
        \label{fig:trust-outcomes-track-1-no-prior}
    \end{subfigure}
    \begin{subfigure}[t]{0.24\linewidth}
        \centering
        \includegraphics[width=\linewidth]{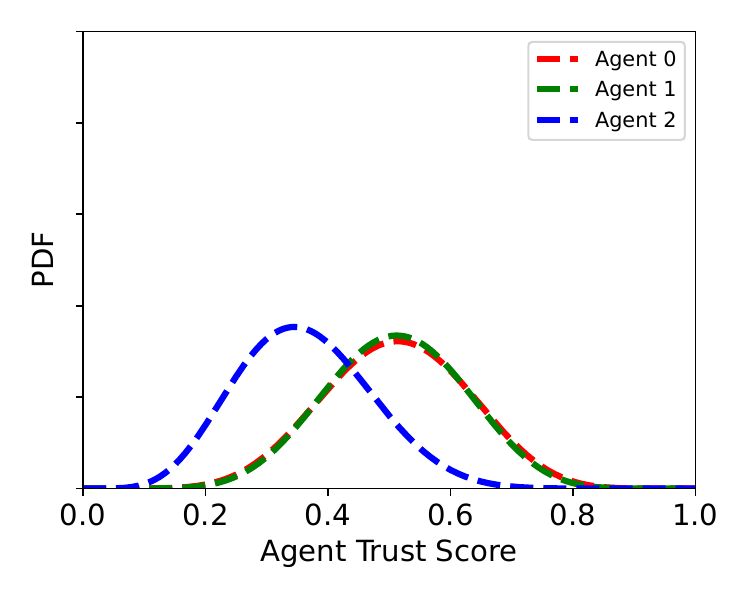}
        \caption{C1: agent trust, uninf. prior}
        \label{fig:trust-outcomes-agent-1-no-prior}
    \end{subfigure}
    %
    \begin{subfigure}[t]{0.24\linewidth}
        \centering
        \includegraphics[width=\linewidth]{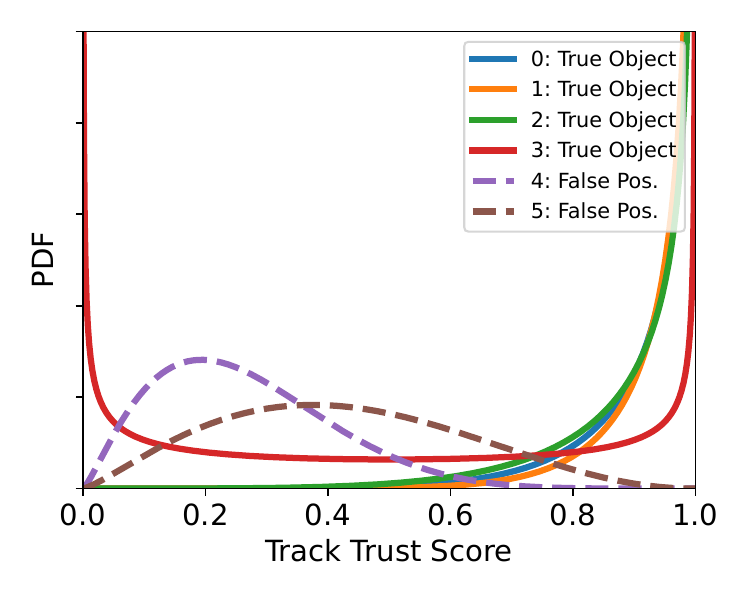}
        \caption{C2: track trust, uninf. prior}
        \label{fig:trust-outcomes-track-2-no-prior}
    \end{subfigure}
    \begin{subfigure}[t]{0.24\linewidth}
        \centering
        \includegraphics[width=\linewidth]{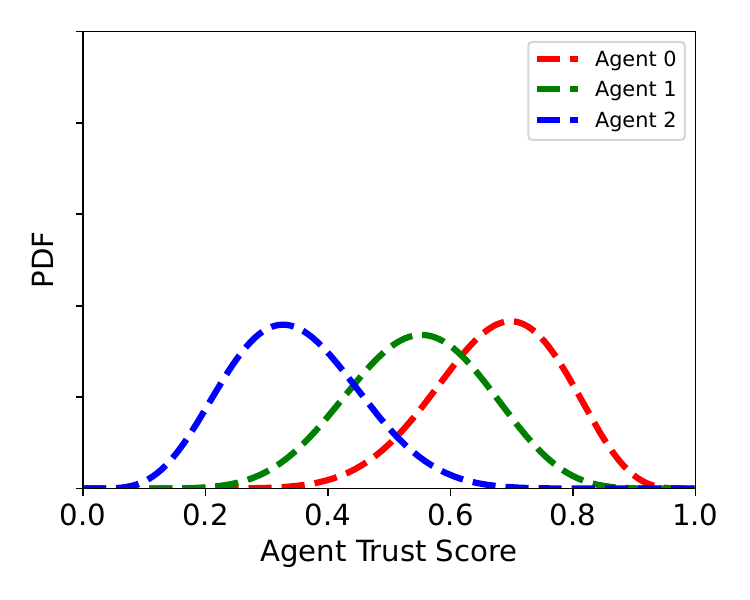}
        \caption{C2: agent trust, uninf. prior}
        \label{fig:trust-outcomes-agent-2-no-prior}
    \end{subfigure}
    %
    \begin{subfigure}[t]{0.24\linewidth}
        \centering
        \includegraphics[width=\linewidth]{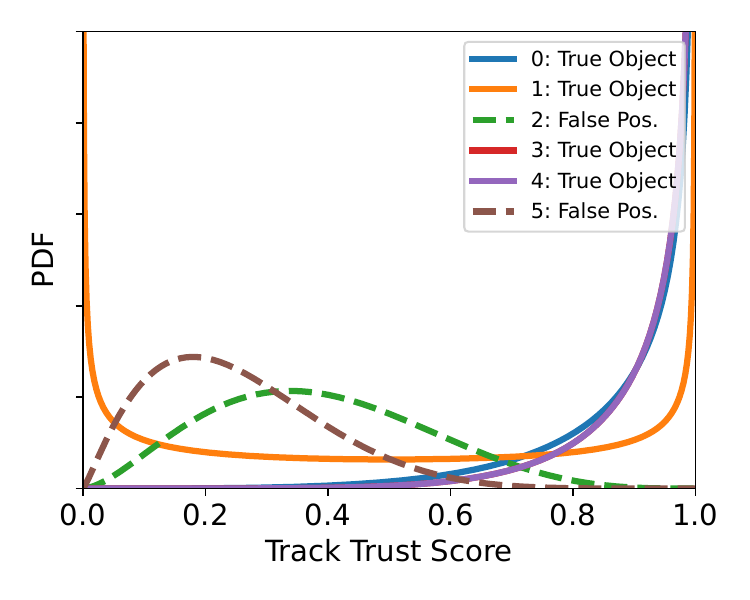}
        \caption{C1: track trust, strong prior}
        \label{fig:trust-outcomes-track-1-prior}
    \end{subfigure}
    \begin{subfigure}[t]{0.24\linewidth}
        \centering
        \includegraphics[width=\linewidth]{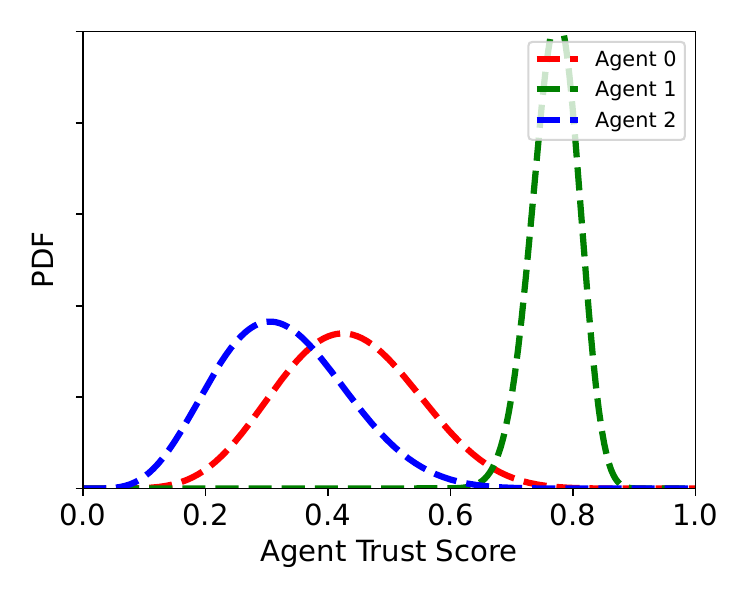}
        \caption{C1: agent trust, strong prior}
        \label{fig:trust-outcomes-agent-1-prior}
    \end{subfigure}
    %
    \begin{subfigure}[t]{0.24\linewidth}
        \centering
        \includegraphics[width=\linewidth]{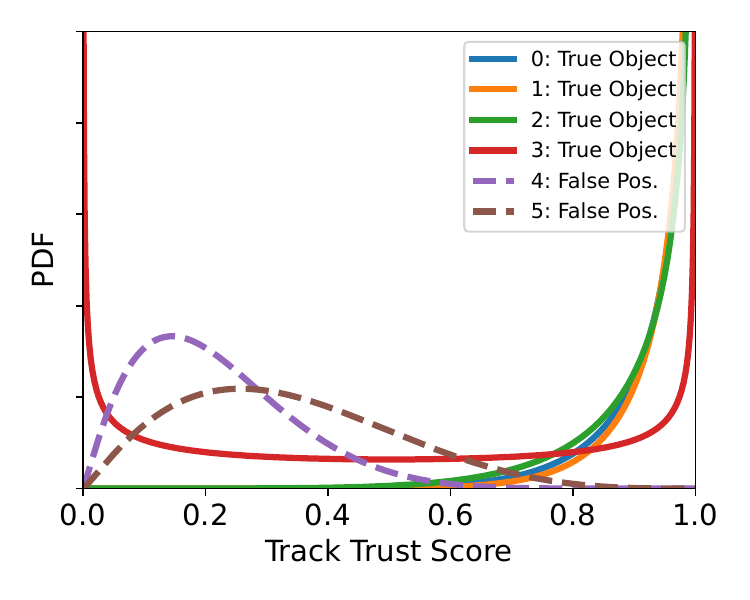}
        \caption{C2: track trust, strong prior}
        \label{fig:trust-outcomes-track-2-prior}
    \end{subfigure}
    \begin{subfigure}[t]{0.24\linewidth}
        \centering
        \includegraphics[width=\linewidth]{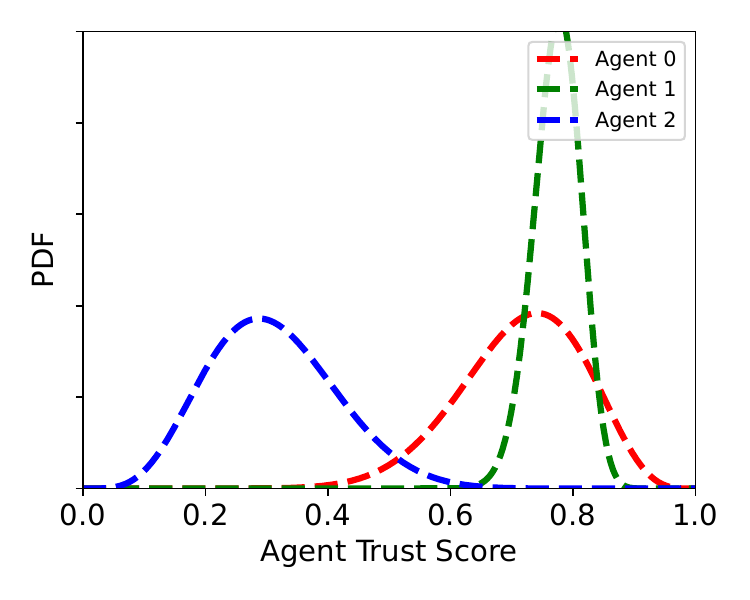}
        \caption{C2: agent trust, strong prior}
        \label{fig:trust-outcomes-agent-2-prior}
    \end{subfigure}
    
    \caption{(a,b,e,f) Case 1 with Agents 0, 2 providing single FP each: (a, b) uninformative priors; difficult to ascertain agent trust due to ambiguity in FP determination with limited numbers of observers. (e, f) Strong Agent 1 prior; clear determination of trusted/distrusted tracks and correspondingly clear agent trust obtained with strong prior \emph{only} on Agent 1 being trusted. (c,d,g,h) Case 2 with Agent 2 providing multiple FPs: (c, d) easier to identify malicious agent compared to Case 1 due to multiple FPs. (g,h) Strong prior on Agent 1 aids trust identification.}
    \label{fig:trust-outcomes}
\end{figure*}

%% file: 6-conclusion.tex
\section{Conclusion}

Track scoring in MTT is provably vulnerable to adversarial attacks even when the number of benign agents significantly outnumbers the adversaries. To improve the security-awareness of MTT, we establish a Bayesian model that estimates the trust of agents and MTT's tracks by mapping the sensing inputs to a real-valued trust pseudomeasurement. Our trust estimation algorithm handles uncertain measurements and provides a probability distribution over the trust based on Bayesian updating. Trust estimation is capable of detecting and identifying adversarial false positive tracks while confirming true tracks as trusted entities in many cases.